\pdfoutput=1

\documentclass[11pt]{article}

\usepackage[final]{coling}
\usepackage{multirow} 
\usepackage{times}
\usepackage{latexsym}
\usepackage{xcolor}
\usepackage{subcaption}

\usepackage[T1]{fontenc}

\usepackage[utf8]{inputenc}

\usepackage{microtype}

\usepackage{inconsolata}

\usepackage{graphicx}

\title{Make Satire Boring Again: Reducing Stylistic Bias of Satirical Corpus by Utilizing Generative LLMs}

\author{
  \textbf{Asli Umay Ozturk\textsuperscript{1}},
  \textbf{Recep Firat Cekinel\textsuperscript{1}},
  \textbf{Pinar Karagoz\textsuperscript{1}}
\\
\\
\textsuperscript{1} Department of Computer Engineering \\
Middle East Technical University (METU)\\
\{auozturk,rfcekinel,karagoz\}@ceng.metu.edu.tr
\\
  \small{
    \textbf{Correspondence:} \href{mailto:auozturk@ceng.metu.edu.tr}{auozturk@ceng.metu.edu.tr}
  }
}

\begin{document}
\maketitle
\begin{abstract}
Satire detection is essential for accurately extracting opinions from textual data and combating misinformation online. However, the lack of diverse corpora for satire leads to the problem of stylistic bias which impacts the models' detection performances.
This study proposes a debiasing approach for satire detection, focusing on reducing biases in training data by utilizing generative large language models. 
The approach is evaluated in both cross-domain (irony detection) and cross-lingual (English) settings. Results show that the debiasing method enhances the robustness and generalizability of the models for satire and irony detection tasks in Turkish and English. However, its impact on causal language models, such as Llama-3.1, is limited. Additionally, this work curates and presents the Turkish Satirical News Dataset with detailed human annotations, with case studies on classification, debiasing, and explainability.
\end{abstract}

\section{Introduction}

\begin{figure*}[ht!]
\centering
\includegraphics[width=0.8\linewidth]{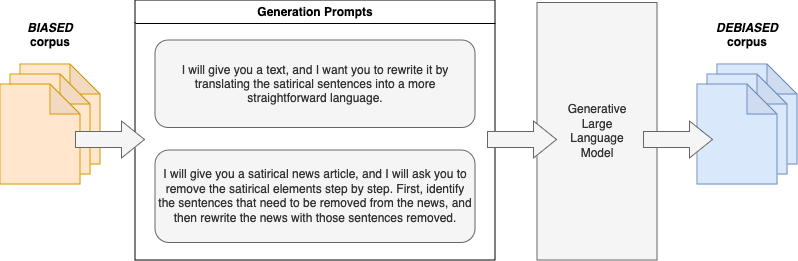}
\caption{The proposed debiasing pipeline}
\label{fig:pipeline}
\vspace{-4mm}
\end{figure*}

There are no universally agreed definitions for satire, sarcasm, and irony in NLP literature. However, Cambridge Dictionary\footnote{\url{https://dictionary.cambridge.org/dictionary/english}} defines \textbf{satire} as a way of \textit{criticizing} people or ideas in a \textit{humorous} way, especially in order to make a political point. Whereas \textbf{sarcasm} is defined as the use of remarks that clearly \textit{mean the opposite} of what they say, made in order to hurt someone's feelings or to \textit{criticize} something in a \textit{humorous} way. Moreover, \textbf{irony} is related to the use of words that are the \textit{opposite of what you mean}, as a way of being \textit{funny}. The overlaps in these definitions cause different studies to use these terms interchangeably, and borrow ideas from other studies \cite{barbieriModelling2014,vanheeExploring2016,vanheeSemeval20182018,carvalhoSituational2020}. 

At first glance, it might not be very apparent why detecting satirical content is important. However, with the increased usage of social media, the primary source of news and information for many people has become the shared news articles in their social media feeds. Even though this makes the information more accessible, it can also cause misinformation to spread at fast rates \cite{allcott2017social, aimeur2023fake}. It is not uncommon for regular social media users to take fake or satirical content as the truth \cite{wuMisinformation2019}, which is specifically problematic when it comes to news content. Hence, satire detection can offer a solution to this misinformation problem since automated detection of satirical content can be used to create automated warnings that inform social media users about the reliability of a piece of information. 

For the last couple of years, with the rise of LLMs, an improvement in the performance of NLP tasks has been seen in the literature \cite{li2022survey}. Even though recent studies have demonstrated high performances for satire, sarcasm, and irony classification tasks using multilingual LLMs, it remains unclear whether these concepts are represented similarly across different languages and domains \cite{ortega2023cross, maladry2023fine}. 

Moreover, focusing on a data-centric approach when training LLMs also raises problems since it is not easy to find a diverse set of resources for low-resourced languages \cite{dougru2018parallel,hangya2022improving,acikgoz2024bridging} or specific tasks. For example, annotating data that can be labeled as satirical, sarcastic, or ironic requires extensive human labor. Instead, automatic data collection processes may be employed, and the data would be collected from a limited number of sources that are already known to belong to the target label. As a result, this creates stylistically unbalanced corpora to be used to train and fine-tune LLMs. This may result in a bias or misalignment in the model \cite{xu2024survey}. In other words, potential stylistic bias in the curated dataset impacts the robustness of the models that use these datasets during training and fine-tuning.

This work aims to reduce the effect of stylistic bias stemming from a single-sourced satirical corpus, proposing a debiasing method that utilizes generative LLMs to reduce the stylistic bias of the instances in the biased corpus.

The proposed method works by generating satirical texts that are stylistically more natural, making the generated corpus more parallel to the non-satirical corpus. This method is demonstrated on a curated dataset for satirical news detection in Turkish. Our contributions can be summarized as follows: 

\begin{itemize}
    \item Curating the \emph{Turkish Satirical News Dataset} with human annotations, and analyzing its stylistic bias and usability\footnote{\url{https://github.com/auotomaton/satiretr}},
    \item Proposing a \emph{debiasing pipeline (Figure \ref{fig:pipeline})} for combating stylistic bias and improving model generalizability,
    \item Analyzing the cross-lingual and cross-domain performance of Turkish satire detection model for irony and English.
\end{itemize}

\section{Related Work}%
\label{chp:2}

Starting in the early 2010s, the related literature began to focus on the problems of binary detection for satire, sarcasm, and irony. Earlier works generally utilize traditional supervised learning methods such as Support Vector Machine (SVM) or Naive Bayes (NB) based classifiers and propose different feature extraction methods for different languages and tasks \cite{buschmeierImpact2014,barbieriModelling2014,vanheeMonday2016,pamungkasNonDicevoSulSerio2018,balogluAssessment2019,ozturk2021ironytr,onan2020satire}. Another approach explored in the earlier studies is utilizing neural network based architectures such as LSTM (Long-Short Term Mermory) networks \cite{wuTHUNGN2018,zhangIrony2019}. Later works started to utilize transformer architectures such as BERT (Bidirectional Encoder Representations from Transformers) \cite{devlinBERT2019} and reported improved results \cite{buyukbasExplainability2021}.  

In the last couple of years, newer studies have focused more and more on multimodal approaches and LLM-based models. In their study,  \citet{tomas2023transformer} explore the irony detection performance of transformer-based models with both textual and visual inputs. On the other hand, \citet{lin2024augmenting} combines transformer-based models with prompt engineering to improve the irony detection performance, specifically focusing on different features of the text. 

A recurring problem in the literature for the aforementioned tasks is the lack of labeled data and openly available datasets. There are curated datasets for irony detection \cite{vanheeExploring2016} and sarcastic news detection \cite{barbieriModelling2014} but they are mostly in English. Even though some other datasets curated for other languages do exist \cite{ortega-buenoOverview2019, xiangCiron2020, ozturk2021ironytr, ghanemIrony2020, joshiAutomatic2017}, they are much smaller than the available English corpus for irony, sarcasm, and satire. 

Literature on irony, satire, and sarcasm detection also includes studies utilizing explainable artificial intelligence (AI) and interpretable machine learning (ML) methods \cite{buyukbasExplainability2021} such as LIME (Local Interpretable Model-Agnostic Explanations) \cite{lime} and SHAP (SHapley Additive exPlanations) \cite{shap} or proposing task-based structures to understand model decisions \cite{shuDefend2019}. These studies aim to analyze the model decisions to improve the performance, fairness, and generalizability of the model, as well as reduce the bias of the model. 

With the advancements in generative LLMs, there have been studies utilizing them to augment training data or synthetically generate data from scratch, to overcome the shortness of training data available for a diverse set of tasks \cite{hangya2022improving,long2024llms}. 

A couple of works have tried to utilize this data generation approach to overcome biases in datasets. \citet{qian2021counterfactual} focus on dataset bias and propose a framework for debiasing using counterfactual inference. They show that their approach improves the effectiveness, generalisability, and fairness of the classifier. In another study, \citet{schlicht2024pitfalls} utilize conversational LLMs to reduce textual bias in news articles. Their findings show that even though they are compelling in some cases, they tend to leave out vital and contextual information during the debiasing process.

Differing from the existing works, this work explores the effects of synthetically generated data on debiasing binary classifiers trained on low-resource languages, focusing on satire detection in Turkish. We propose a debiasing pipeline that aims to improve the cross-lingual and cross-domain performance of a model trained on a stylistically biased dataset.

\section{Dataset}
\label{sec:dataset}

One contribution of this study is to curate an open dataset for satire and satirical news detection in Turkish. Utilizing the satirical news publication Zaytung\footnote{\url{https://zaytung.com/}} and Turkish news agency Anadolu Agency (AA)\footnote{\url{https://www.aa.com.tr/}}, \emph{Turkish Satirical News Dataset} is curated.

\subsection{Curation of Turkish Satirical News Dataset}

As a source of satirical news articles, the Turkish satirical news publication Zaytung is used. By crawling the Zaytung website archive, 2825 satirical articles are collected with \textit{timestamp}, \textit{title}, \textit{body} and \textit{header image} information. To improve the representative nature of the dataset, articles dated before 2014 are discarded. 

As a source of non-satirical news articles, AA archives are crawled between the dates 2022-2023, and 4781 articles are collected with \textit{id}, \textit{title}, \textit{subtitle}, \textit{author}, \textit{date}, \textit{category}, \textit{header image}, \textit{city}, \textit{body} and \textit{tag} information. 

The final dataset includes 2202 \texttt{SATIRICAL} and 4781 \texttt{NON-SATIRICAL} articles. Additionally, 40 of the \texttt{SATIRICAL} instances have word-by-word human annotations aiming to capture the effect of each word on the satirical meaning of the article. The annotation process and a case study comparing the annotations with model explanations are presented in Appendix \ref{sec:app_annotate} and \ref{sec:app_shapselin}.

The curated dataset is available publicly on GitHub\footnote{\url{https://github.com/auotomaton/satiretr}} for other researchers' use. The GitHub repository also includes the debiased articles generated from the Zaytung corpus, described in Section \ref{sec:method}. All data is scraped from publically available sources and Zaytung has specifically been reached for consent to use their articles.

\subsection{Bias Analysis}

Since both the \texttt{SATIRICAL} and the \texttt{NON-SATIRICAL} corpora are taken from a single publication each, it might be expected to see a stylistic and statistical bias that results in an easily separable dataset. This may result in classifiers that are trained using such datasets to become biased for the style elements of the corpus instead of identifying satire/non-satire indications of the text. One simple way to observe this is by conducting statistical analysis on both the \texttt{SATIRICAL} and \texttt{NON-SATIRICAL} corpora.

\subsubsection{Average Word and Sentence Count}

A primary statistical analysis is performed and reported in Table \ref{tab:basic-stats} to better understand the data instances. It can be seen that \texttt{SATIRICAL} corpus has an average of 329 words per instance and 44 sentences per instance. On the other hand, \texttt{NON-SATIRICAL} corpus has an average of 313 words per instance and 43 sentences per instance. Even though the numbers are close, on average, we see that the \texttt{SATIRICAL} corpus has more words per sentence.

\begin{table} [th!]
  \centering
  \resizebox{\linewidth}{!}{
  \begin{tabular}{lll}
    \hline
    \textbf{Statistic} & \texttt{SATIRICAL} & \texttt{NON-SATIRICAL} \\
    \hline
    avg \#of words & 329 & 313 \\
    \hline
    avg \#of sentences & 44 & 43 \\
    \hline
  \end{tabular}
  }
  \caption{Statistics of the \emph{Turkish Satirical News Dataset} by corpus}
  \label{tab:basic-stats}
  \vspace{-4mm}
\end{table}

\subsubsection{Top 10 Words}

\begin{table}[th!]
  \centering
\resizebox{\linewidth}{!}{%
  \begin{tabular}{ll}
    \hline
    \textbf{Label} & \textbf{Top 10 Words} \\
    \hline
    \texttt{SATIRICAL}  & \textit{almak (take), bir (one/a),}\\
    & \textit{demek (say), etmek (make),} \\
    & \textit{gelmek (come), iş (work/job),}\\
    & \textit{olarak (being), vermek (give),} \\
    & \textit{türkiye (Turkiye), yapmak (do)} \\
    \hline
    \texttt{NON-SATIRICAL} & \textit{ülke (country), yıl (year),} \\
    & \textit{açıklama (explanation),}\\
    & \textit{ifade (expression),}\\
    & \textit{fotoğraf (photograph), } \\
    & \textit{spor (sport), bölge (region), }\\
    & \textit{başkan (president),}\\
    & \textit{konu (issue), çalışmak (work)}\\
     
  \end{tabular}
  }\caption{Statistics of the \emph{Turkish Satirical News Dataset} by Top 10 Words}
  \label{tab:top10}
  \vspace{-5mm}
\end{table}

To have a general idea about the content of the news belonging to both the \texttt{SATIRICAL} and \texttt{NON-SATIRICAL} corpora, top-10 terms are extracted per label by TF-IDF (Term Frequency - Inverse Document Frequency) scoring. These terms are shown in Table \ref{tab:top10}. It is visible that the top 10 words for the \texttt{SATIRICAL} and \texttt{NON-SATIRICAL} corpora do not have any words in common. This also follows the idea that the tones of the two corpora are different.

\section{The Proposed Debiasing Method}
\label{sec:method}

Bias introduced from the style of the source of the corpus is a serious concern that is hard to eliminate without extensive human annotation. This is specifically prevalent in fairly low-resource languages \cite{shen2024language} and text qualities such as satirical, sarcastic, or ironic meaning \cite{maladry2023fine,ortega2023cross}. Using a single source as a corpus makes the models more prone to bias. Hence, this study proposes the \emph{debiasing pipeline} utilizing LLMs for generating less biased counterparts of texts.

\subsection{Pipeline Design}

The \emph{debiasing pipeline} proposed in this work utilizes prompt engineering and synthetic data generation to remove the effect of the bias coming from the heavily stylistic language of the \texttt{SATIRICAL} corpus. The proposed pipeline is created to improve the usability of the curated dataset, however, we believe the pipeline can be generalized for any biased dataset by adapting the generation prompts according to the task and the bias.

The \emph{debiasing pipeline}, summarised in Figure \ref{fig:pipeline}, generates a new set of data to be used to replace the training instances from the biased corpus. In the scope of this work, the stylistically biased data in the train set is the \texttt{SATIRICAL} class coming from the scraped Zaytung corpus.

\subsection{Generating "Debiased" Articles Using Prompt Engineering}

This subsection goes over the prompt engineering and debiased instance generation using sample articles translated so that it will be easier for readers to follow. The original Turkish articles and Turkish prompts are available in Appendix \ref{sec:app_prompts}. In this study, the data generation is done via the GPT web interface, ChatGPT\footnote{\url{https://chatgpt.com/}}. However, the generation pipeline can easily be automatized by using the OpenAI API to fully automate the debiasing process. The following two prompts are used to generate stylistically less biased articles:
\vspace{-1mm}
\paragraph{Prompt 1:} \emph{"I will give you a satirical news article, and I will ask you to remove the satirical elements step by step. First, identify the sentences that need to be removed from the news, and then rewrite the news with those sentences removed. \\ Article text:"}
\vspace{-2mm}
\paragraph{Prompt 2:} \emph{"I will give you a text, and I want you to rewrite it by translating the satirical sentences into a more straightforward language. \\ Article text:"} 

The first prompt, Prompt 1, asks the generative LLM to identify the satirical elements in the text and rewrite the article by excluding them. On the other hand, Prompt 2 asks the generative LLM to directly rewrite the article by modifying the sentences with satirical meaning to have a more straightforward language. Both prompts put the generative LLM to test in terms of its language understanding. 

\begin{figure*}[!ht]
    \centering
    \begin{subfigure}[b]{0.98\textwidth}
        \centering
        \fbox{\parbox{\textwidth}{%
        {\scriptsize
\textbf{NASA Announces Termination of Mars Program Due to Photos Not Reaching Expected Like Counts…}

The American Space Agency NASA is facing troubled days due to the photos sent by the Perseverance rover following its landing on Mars.

NASA’s one-ton Rover-type spacecraft, Perseverance, landed successfully on Mars' Jezero Crater at 15:55 Eastern Time on Thursday after a journey of approximately 7 months. However, 24 hours after the landing, when NASA officials shared the first photos from the red planet on their Instagram account, they expressed surprise and disappointment over the comments they received. The photos, which were expected to garner at least 10 million likes, only managed to accumulate 2 million, leading to significant disillusionment within the community.

Comments such as "Did you send an iron to Mars to take these photos?", "You’d think they’d put a decent camera on it", and "There was no need to go all the way to Mars for this, we could have sorted it out in Yozgat city center" were particularly hurtful, according to James Watzin, Director of NASA’s Mars Program. Watzin admitted, "What’s truly upsetting is that our followers are absolutely right. Honestly, we weren’t satisfied either. If it wasn’t our own rover, we wouldn’t have liked those photos. It’s just not appealing…"

Watzin pointed out that the photos failed to receive as many likes as a makeup-free selfie by Cardi B, noting, "If we don’t receive something like a group of Martian mating photos in the next few days, it seems impossible to reach the number of likes needed to cover the program’s costs. The effort we put in seems to have missed its mark. Why did this happen? Did we choose the wrong cameras? Is Mars itself not photogenic? These are questions that need answering." He announced that the Mars program will be put on hold until the source of the issue is understood.

}
        }}
        \vspace{-2mm}
        \caption{The original article, translated to English}
        \label{fig:orig-text}
    \end{subfigure}
    \vfill
    \begin{subfigure}[b]{0.98\textwidth}
        \centering
        \fbox{\parbox{\textwidth}{%
        {\scriptsize

The American Space and Aeronautics Agency NASA is going through difficult days due to the photos sent after the Perseverance rover successfully landed on Mars. Contrary to expectations, NASA officials encountered low engagement on social media following the successful landing of the Mars program. Program director James Watzin admitted that the lack of expected interest has led to disappointment and that the program has been deemed a failure. The Mars program has been paused to understand the source of the problem.

        }}}
        \vspace{-2mm}
        \caption{Generated article, translated to English}
        \label{fig:generated-text1}
    \end{subfigure}
    \vspace{-2mm}
    \caption{Debiasing of Sample Article (1) with Prompt 1}
    \label{fig:comparison-1}
\end{figure*}

First, consider Sample Article 1 and the article generated from it using Prompt 1, shown in Figure \ref{fig:comparison-1}. Even though the prompt clearly asks for the removal of satirical elements, the satire created by the fake social media expectation narrative is still present in the generated article. However, we see that the style of the writing is less exaggerated. This makes Prompt 1 a good candidate for generation in some cases.

\begin{figure*}
    \centering
    \begin{subfigure}[b]{0.98\textwidth}
        \centering
        \fbox{\parbox{\textwidth}{%
        {\scriptsize
\textbf{The Crisis in LUNA is Growing: It Has Been Revealed to Have Lost Value Even Against the Turkish Lira...}

Following a 99.8\% loss in value, where Terra Luna Coin dropped from \$64 to \$0.20 within two days, bad news continues to emerge. After this dramatic drop that shook the cryptocurrency market, Do Kwon, CEO of Terraform Labs, released a statement today. He noted that their investigation revealed Luna had lost value not only against the US dollar but also against the Turkish Lira. Kwon said, "Everything we thought couldn’t happen in the last two days has happened. I am very sorry."

Kwon explained that the realization came during a meeting he held to boost the team’s morale, saying, “I gathered the team to give a motivational speech along the lines of ‘Look, it’s not so bad. At least we are still gaining value in TL terms.’ I hadn’t felt the need to check if we were worse off compared to TL. That was the real mistake.” He shared that the painful truth became apparent when they opened the graph on the screen.

“At that moment, half the team left the building. We still can’t reach them. The remaining team members are just looking around the office to see if there’s anything left to sell,” Kwon described the current situation at Terra Labs. The experienced CEO also mentioned that they expect the depreciation of the TL to eventually surpass Luna, whether due to some absurd move by Turkey or the Central Bank of the Republic of Turkey depleting its remaining foreign currency reserves. He added that he has not neglected to give optimistic messages for the future.

}
        }}
        \vspace{-2mm}
        \caption{The original satirical article, translated to English}
        \label{fig:orig-text2}
    \end{subfigure}
    \vfill
    \begin{subfigure}[b]{0.98\textwidth}
        \centering
        \fbox{\parbox{\textwidth}{%
        {\scriptsize
        
Terra Luna Coin has experienced a significant loss in value in recent days. Terraform Labs CEO Do Kwon stated that their investigation revealed that Luna has lost value not only against the US dollar but also against the Turkish Lira (TL). This situation has caused a morale decline among company employees. Kwon expressed that in the short term, they expect the depreciation of the TL to surpass that of Luna.

        }}}
        \vspace{-2mm}
        \caption{Generated article with Prompt 1, translated to English}
        \label{fig:generated-text3}
    \end{subfigure}
    \begin{subfigure}[b]{0.98\textwidth}
        \centering
        \fbox{\parbox{\textwidth}{%
        {\scriptsize

In the past two days, Terra Luna Coin has experienced a 99.8\% loss in value, dropping from \$64 to \$0.2, and bad news continues to emerge. Following this major drop, Terraform Labs CEO Do Kwon stated in a recent announcement that they have found evidence that Luna has lost value not only against the US dollar but also against the Turkish Lira. Kwon said, "Everything we didn’t expect to happen in these last two days happened. I’m very sorry."

Kwon mentioned that they realized the situation during a meeting he organized to boost the team’s morale. "I wanted to tell the team, 'At least we’re gaining value against the TL,' but I hadn’t checked the situation. That was the real mistake," he said. He also mentioned that they faced the harsh reality when they opened the chart.

"During the meeting, half of the team left the building, and we still can't reach them. The rest are just looking for things to sell in the office," Kwon said. He also noted that in the short term, the loss of value of the TL might surpass that of Luna, and he did not forget to provide an optimistic message for the future.

        }}}
        \vspace{-2mm}
        \caption{Generated article with Prompt 2, translated to English}
        \label{fig:generated-text4}
    \end{subfigure}
    \vspace{-2mm}
    \caption{Debiasing of Sample Article (2) with Prompt 1 and Prompt 2}
    \label{fig:comparison-2}
    \vspace{-2mm}
\end{figure*}

On the other hand, consider Sample Article 2 and the articles generated from it using both Prompt 1 and Prompt 2, shown in Figure \ref{fig:comparison-2}. It can be seen that Prompt 1  removes most of the sentences contributing to the satire in the text which significantly reduces the satirical meaning of the overall article. Without background knowledge about the financial situation revolving around Turkish Lira (TL), the article becomes vaguely satirical or even non-satirical to the uninformed reader. However, the generated article with Prompt 2 can still be classified as satirical easily, while stripped from some of the more dramatic and stylistically strong phrases (e.g. \emph{"...absurd move by Turkey or the Central Bank of the Republic of Turkey..."} etc.). This makes Prompt 2 also a good candidate for generation.

Another successful example that shows an original-generated article pair generated using Prompt 2 can be seen in Appendix \ref{sec:app_prompts}. With these explorations done on the prompts and their performances, both prompts are used during the generation process of 200 debiased articles.

\vspace{-2mm}
\subsection{Comparing Generated and Original Articles}
\vspace{-1mm}

The proposed debiasing method is expected to keep the satirical value of the articles while making them \\
\noindent stylistically less biased. To check if this is the case, two analyses are conducted. 

Firstly, the generated articles are checked manually (by one annotator) to see if they still can be classified as \texttt{SATIRICAL}. Additionally, they are analyzed to see if the original context of the article is still understandable or if some of the context (such as the events, relationships between people, and such) has been lost during the generation process. It is seen that: 
\vspace{-1mm}
\begin{itemize}
    \item Out of the 200 articles, 29 of them can be labeled as \texttt{NON-SATIRICAL} by an unsuspecting reader with not enough knowledge of the Turkish political landscape. \footnote{Authors are aware that this definition of such a reader is highly subjective. This statistic is obtained to have a surface-level understanding of the quality of the generated content.}
    \item Out of the aforementioned 29 articles, all 29 of them differ from their original counterparts with a loss of important contextual information, such as the total erasure of events or people occurring during the generation process.
    \item Out of the aforementioned 29 articles, 28 of them are generated with Prompt 1. Since Prompt 1 explicitly asks the model to remove satirical sentences, it is understandable to see a major loss in contextual information. This also points to Prompt 2 as being a better option for article generation.
\end{itemize}

As a second way of verifying whether the content of the articles in the corpus is maintained, the BERTScore \cite{zhang2019bertscore} evaluation metric, which calculates the pairwise semantic similarity of tokens in the given pair of sentences using BERT's contextual embeddings, is employed. We used the BERTurk model for extracting contextual embeddings and the cosine similarity of the original sarcastic news and the debiased counterparts is 0.6852 (in terms of F1-binary). 
This similarity score implies that the debiased text mostly retains the original content. 

\begin{figure*}[!ht]
\centering
\includegraphics[width=0.75\linewidth]{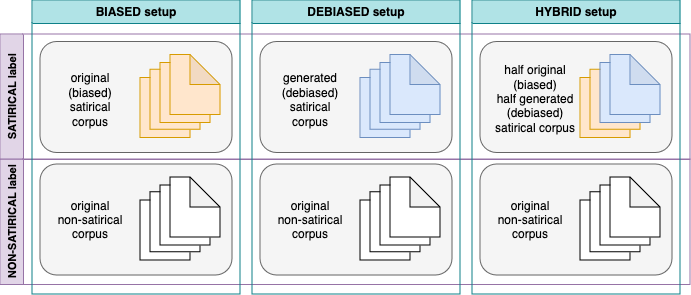}
\caption{Training setups in the experiments}
\label{fig:setups}
\vspace{-4mm}
\end{figure*}

\section{Experiments}

The experiments were conducted on 4 Nvidia A6000 GPUs. We employed three training strategies to evaluate the proposed debiasing pipeline's effectiveness. 

\subsection{Models and Parameters}
We employed multilingual BERT \cite{devlin-etal-2019-bert}, BERTurk \cite{stefan_schweter_2020_3770924}, XLM-Roberta-large \cite{conneau-etal-2020-unsupervised} and Llama-3.1-8B-Instruct \cite{dubey2024llama} models for conducting the experiments. These models were mainly selected because they were also pre-trained in Turkish. Llama-3.1-8B is a generative causal language model and we used sampling-based decoding to make predictions. As a result, for some test cases, the model's inferences do not comply with the expected labels and we reported the rate of nonresponses in Table \ref{tab:nonresponse}. The samples where the model failed to return any labels were excluded during the evaluation of the models on the test sets.

\begin{table}[th!]
\resizebox{\linewidth}{!}{%
\begin{tabular}{l|rrr}
\multicolumn{1}{c}{\textbf{Dataset}} & \multicolumn{1}{c}{\textbf{COMBINED}} & \multicolumn{1}{c}{\textbf{BIASED}} & \multicolumn{1}{c}{\textbf{DEBIASED}} \\
\hline
Zaytung                              & 0.009                                 & 0.002                               & 0.000                                 \\
Onion                                & 0.013                                 & 0.207                               & 0.030                                 \\
IronyTR                              & 0.057                                 & 0.012                               & 0.002                                
\end{tabular}
}
\caption{Rate of nonresponses of Llama-3.1-8B-Instruct model}
\label{tab:nonresponse}
\end{table}

During the training, we randomly selected 10\% of the training data for validation. A grid search was conducted to explore the following hyperparameters: \textit{learning rate} {2e-5, 5e-5, 1e-4} and \textit{batch size} {8, 16, 32}. The best results, as determined through this process, are reported in Section \ref{sec:results}. All models were trained on the training datasets for two epochs.

For the Llama-3.1 model, we set the sequence length to 2048, while for other models, the maximum sequence length was set to 512. Additionally, we employed QLoRA \cite{dettmers2024qlora} for training the Llama model, with the LoRA rank and LoRA alpha set to 32 and the LoRA dropout to 0.05. We used the Adam optimizer with a cosine scheduler and trained the models with fp16 precision.

\subsection{Training Setups and Test Datasets}

The models were trained on three different setups as follows (illustrated in Figure \ref{fig:setups}): 

\paragraph{BIASED:} 200 instances from the \texttt{SATIRICAL} corpus and 200 instances from the \texttt{NON-SATIRICAL} corpus in \emph{Turkish Satirical News Dataset} (detailed in Section \ref{sec:dataset}) are selected. 

\paragraph{DEBIASED:} Same sets of instances as the \emph{BIASED} setup are selected first, then the selected \texttt{SATIRICAL} instances are passed through the proposed \emph{debiasing pipeline}. Final \emph{DEBIASED} setup includes 200 instances from the \texttt{SATIRICAL} corpora and 200 instances from the \texttt{NON-SATIRICAL} corpora, where the \texttt{SATIRICAL} instances are debiased stylistically.

\paragraph{HYBRID:} To have an intermediate setup between \emph{BIASED} and \emph{DEBIASED}, the same 200 \texttt{SATIRICAL} instances are selected, but only 100 of them are passed through the \emph{debiasing pipeline}. Hence, the final \emph{HYBRID} setup consists of 200 instances from \texttt{NON-SATIRICAL} corpora, 100 instances from original \texttt{SATIRICAL} corpora, and 100 instances from debiased \texttt{SATIRICAL} corpora.

After training, the models were evaluated in same-domain, cross-domain, and cross-lingual settings using the following test datasets:

\paragraph{Zaytung + AA:} All instances from the \emph{Turkish Satirical News Dataset} that are not used in the training sets are included in this test set. Hence, this set consists of 4581 instances from the \texttt{NON-SATIRICAL} corpora and 2002 instances from the \texttt{SATIRICAL} corpora of the \emph{Turkish Satirical News Dataset}.

\paragraph{The Onion + HuffPost:} A fairly balanced set of 29000 English news article headlines from the American satirical news website the Onion\footnote{\url{https://theonion.com/}} and HuffPost\footnote{\url{https://www.huffpost.com/}}, taken from the openly available News Headlines Dataset For Sarcasm Detection\footnote{\url{https://www.kaggle.com/datasets/rmisra/news-headlines-dataset-for-sarcasm-detection/data}}.

\paragraph{IronyTR:} IronyTR \cite{ozturk2021ironytr} is a Turkish social media irony detection dataset annotated by humans. It contains 300 Turkish ironic short texts and 300 non-ironic social media posts. This dataset is used to evaluate the cross-domain performance of the models trained under different settings.

\subsection{Results}
\label{sec:results}

\begin{table*}[th!]
\centering
\resizebox{\linewidth}{!}{
\begin{tabular}{l|rrrr|rrrr|rrrr}
\multicolumn{1}{c}{}  & \multicolumn{4}{c}{\textbf{BIASED}} & \multicolumn{4}{c}{\textbf{DEBIASED}} & \multicolumn{4}{c}{\textbf{HYBRID}} \\
\hline
\multicolumn{1}{c}{\textbf{Models}} & \multicolumn{1}{c}{\textbf{accuracy}} & \multicolumn{1}{c}{\textbf{precision}} & \multicolumn{1}{c}{\textbf{recall}} & \multicolumn{1}{c}{\textbf{f1-macro}} & \multicolumn{1}{c}{\textbf{accuracy}} & \multicolumn{1}{c}{\textbf{precision}} & \multicolumn{1}{c}{\textbf{recall}} & \multicolumn{1}{c}{\textbf{f1-macro}} & \multicolumn{1}{c}{\textbf{accuracy}} & \multicolumn{1}{c}{\textbf{precision}} & \multicolumn{1}{c}{\textbf{recall}} & \multicolumn{1}{c}{\textbf{f1-macro}} \\
\hline
berturk & 93.50\% & 93.95\% & 93.50\% & \textbf{92.56\%}    
& 78.74\% & 77.90\% & 78.74\% & 72.26\% \textcolor{orange}{(-20.30\%)}
& 90.52\% & 90.45\% & 90.52\% & 88.41\% \textcolor{orange}{(-4.15\%)}\\
mbert-base & 95.23\% & 95.59\% & 95.23\% & \textbf{94.53\%} 
& 56.11\% & 56.21\% & 56.11\% & 56.01\% \textcolor{orange}{(-38.55\%)}  
& 94.68\% & 94.66\% & 94.68\% & 93.68\% \textcolor{orange}{(-0.85\%)}\\
xlm-roberta large & 97.83\% & 97.86\% & 97.83\% & \textbf{97.39\%}      
& 93.61\% & 93.96\% & 93.61\% & 92.01\% \textcolor{orange}{(-5.38\%)} 
& 96.63\% & 96.73\% & 96.63\% & 95.90\% \textcolor{orange}{(-1.49\%)}\\
llama-3.1-8B & 67.26\% & 84.18\% & 67.26\% & 67.14\%
& 65.28\% & 83.79\% & 65.28\% & 65.21\% \textcolor{orange}{(-1.93\%)} 
& 89.21\% & 91.40\% & 89.21\% & \textbf{88.17\%} \textcolor{blue}{(+21.03\%)}
\end{tabular}
}
\caption{Evaluation on Zaytung + AA dataset}
\label{tab:test-zaytung}
\end{table*}

\begin{table*}[th!]
\centering
\resizebox{\linewidth}{!}{
\begin{tabular}{l|rrrr|rrrr|rrrr}
 \multicolumn{1}{c}{}  & \multicolumn{4}{c}{\textbf{BIASED}} & \multicolumn{4}{c}{\textbf{DEBIASED}} 
    & \multicolumn{4}{c}{\textbf{HYBRID}}   
    \\
\hline
 \multicolumn{1}{c}{\textbf{Models}} & \multicolumn{1}{c}{\textbf{accuracy}} & \multicolumn{1}{c}{\textbf{precision}} & \multicolumn{1}{c}{\textbf{recall}} & \multicolumn{1}{c}{\textbf{f1-macro}} & \multicolumn{1}{c}{\textbf{accuracy}} & \multicolumn{1}{c}{\textbf{precision}} & \multicolumn{1}{c}{\textbf{recall}} & \multicolumn{1}{c}{\textbf{f1-macro}} & \multicolumn{1}{c}{\textbf{accuracy}} & \multicolumn{1}{c}{\textbf{precision}} & \multicolumn{1}{c}{\textbf{recall}} & \multicolumn{1}{c}{\textbf{f1-macro}} \\
\hline
berturk & 48.44\% & 48.09\% & 48.44\% & 47.78\% 
& 52.38\% & 52.68\% & 52.38\% & \textbf{52.37\%} \textcolor{blue}{(+4.59\%)} 
& 50.41\% & 51.39\% & 50.41\% & 49.56\% \textcolor{blue}{(+1.78\%)} \\
mbert-base & 49.29\% & 48.44\% & 49.29\% & 47.34\% 
& 55.02\% & 56.26\% & 55.02\% & \textbf{54.49\%} \textcolor{blue}{(+7.15\%)}
& 56.08\% & 58.26\% & 56.08\% & 49.27\% \textcolor{blue}{(+1.93\%)}\\
xlm-roberta large & 52.45\% & 73.32\% & 52.45\% & 34.58\% 
& 56.53\% & 56.43\% & 56.53\% & 55.09\% \textcolor{blue}{(+20.51\%)}
& 63.36\% & 63.72\% & 63.36\% & \textbf{62.34\%} \textcolor{blue}{(+27.76\%)}\\
llama-3.1-8B & 70.95\% & 71.39\% & 70.95\% & \textbf{70.70\%} 
& 69.54\% & 72.63\% & 69.54\% & 68.81\% \textcolor{orange}{(-2.14\%)}
& 64.87\% & 66.36\% & 64.87\% & 63.23\% \textcolor{orange}{(-7.72\%)}
\end{tabular}
}
\caption{Evaluation on The Onion + HuffPost datasets}
\label{tab:test-onion}
\end{table*}

\begin{table*}[th!]
\centering
\resizebox{\linewidth}{!}{
\begin{tabular}{l|rrrr|rrrr|rrrr}
\multicolumn{1}{c}{}& \multicolumn{4}{c}{\textbf{BIASED}}& \multicolumn{4}{c}{\textbf{DEBIASED}}& \multicolumn{4}{c}{\textbf{HYBRID}}\\
\hline
\multicolumn{1}{c}{\textbf{Models}} & \multicolumn{1}{c}{\textbf{accuracy}} & \multicolumn{1}{c}{\textbf{precision}} & \multicolumn{1}{c}{\textbf{recall}} & \multicolumn{1}{c}{\textbf{f1-macro}} & \multicolumn{1}{c}{\textbf{accuracy}} & \multicolumn{1}{c}{\textbf{precision}} & \multicolumn{1}{c}{\textbf{recall}} & \multicolumn{1}{c}{\textbf{f1-macro}} & \multicolumn{1}{c}{\textbf{accuracy}} & \multicolumn{1}{c}{\textbf{precision}} & \multicolumn{1}{c}{\textbf{recall}} & \multicolumn{1}{c}{\textbf{f1-macro}} \\
\hline
berturk & 53.10\% & 53.28\% & 53.10\% & 52.67\% 
& 59.46\% & 60.81\% & 59.46\% & \textbf{58.01\%} \textcolor{blue}{(+5.34\%)}
& 51.59\% & 51.58\%  & 51.59\% & 51.56\% \textcolor{orange}{(-1.11\%)}\\
mbert-base & 62.48\% & 64.51\% & 62.48\% & \textbf{61.23\%} 
& 56.11\% & 56.21\% & 56.11\% & 56.01\% \textcolor{orange}{(-5.22\%)}
& 58.29\% & 58.41\% & 58.29\% & 58.08\% \textcolor{orange}{(-3.15\%)}\\
xlm-roberta large & 52.93\% & 55.27\% & 52.93\% & 46.43\% 
& 64.99\% & 70.61\% & 64.99\% & 62.56\% \textcolor{blue}{(+16.13\%)} 
& 68.01\% & 68.77\% & 68.01\% & \textbf{67.72\%} \textcolor{blue}{(+21.29\%)}\\
llama-3.1-8B& 64.58\% & 64.59\% & 64.58\% & \textbf{64.55\%}& 
62.58\% & 70.90\% & 62.58\% & 58.69\% \textcolor{orange}{(-5.86\%)}  
& 64.83\% & 64.98\% & 64.83\% & 64.47\% \textcolor{orange}{(-0.08\%)}
\end{tabular}
}
\caption{Evaluation on the IronyTR dataset}
\label{tab:test-ironytr}
\end{table*}

Table \ref{tab:test-zaytung} presents the fine-tuning results of the selected language models on the \textit{BIASED}, \textit{DEBIASED}, and \textit{HYBRID} setups which were evaluated on the Zaytung test set. According to the results, for each model except the Llama-3.1-8B, the \textit{BIASED} setup achieved the highest F1-macro score. Since the training and test sets are from the same domain and the writing styles of the satirical and non-satirical news are significantly different, the outcome is expected. However, the proposed debiasing pipeline significantly reduced the F1-macro score across all language models. This reduction is attributed to the pipeline’s goal of changing the writing style and diminishing the sarcastic tone in the texts, by making it more challenging for the models to differentiate between non-satirical and satirical news.

Secondly, Table \ref{tab:test-onion} presents the cross-lingual evaluation results where the models trained on the Turkish dataset were tested using The Onion dataset. The results show that, except for Llama-3.1-8B, the proposed debiasing approach positively improved the F1-macro scores. In other words, the XLM-RoBERTa model achieved the highest score on the \textit{HYBRID} dataset, which includes debiased instances, while the BERT models performed best on the \textit{DEBIASED} setup.

Finally, Table \ref{tab:test-ironytr} presents the models' performance in a cross-domain setting using the IronyTR dataset which contains ironic and non-ironic social media posts. Since the social media posts are short texts, whereas the training instances are long-form news articles, the models’ performance was significantly lower than the results in Table \ref{tab:test-zaytung}. The proposed debiasing pipeline positively impacted the BERTurk and XLM-RoBERTa models; however, the highest F1-macro scores for multilingual BERT and Llama-3.1-8B were observed in the \textit{BIASED} setup. While the Llama-3.1-8B model achieved its best F1-macro scores on both The Onion and IronyTR datasets using the \textit{BIASED} setup, the scores were very close to those obtained in the other setups. 

\subsection{Discussion}
Following the \emph{DEBIASED} training, the masked language models demonstrated improved robustness in both cross-lingual (see Table \ref{tab:test-onion}) and cross-domain settings (see Table \ref{tab:test-ironytr}). However, for the Llama-3.1-8B model, \emph{BIASED} training achieved the highest score, though the margin compared to other setups was minimal. More specifically in Table \ref{tab:test-onion}, \emph{BIASED} setup outperformed \emph{DEBIASED} setup by 2.14\%, while in Table \ref{tab:test-ironytr}, it outperformed the \emph{HYBRID} setup by 0.08\%. This performance in \emph{BIASED} setups for LLama can be attributed to the model's pretraining knowledge. In other words, articles from The Onion, HuffPost, and instances of IronyTR might have been included in the pretraining data for the Llama model.  Furthermore, Llama significantly outperformed the masked language models on The Onion + HuffPost dataset, further suggesting potential exposure during pretraining. Lastly, on \emph{HYBRID} setup, XLM-RoBERTa outperformed the \emph{BIASED} setup on both cross-lingual and cross-domain evaluations (see Table \ref{tab:test-onion} and Table \ref{tab:test-ironytr}) with a significant margin. This result indicates that combining biased and debiased articles contributes to the model's robustness.

\section{Conclusions}
The problem of satire detection demands a human in the loop by its nature since the labeling process cannot be automated. The only automatization possible is finding a satirical resource (such as Zaytung for Turkish) and assuming all scraped content is satirical by default. Unfortunately, this causes the data to be biased stylistically and trickles down this bias to the model where the model learns to identify the style of the corpus instead of the satire.

This work proposes a debiasing method utilizing LLM-based text generation within ethical limits. We show that generating data that is stylistically neutral to replace the biased data in the training set decreases the model performance significantly and improves the cross-lingual and cross-domain robustness of the model for satire detection in Turkish. However, additional experimentation is needed to see if this method is generalizable as a debiasing method for different language tasks. Yet, the obtained results are promising to demonstrate the applicability of the proposed method.

\section{Limitations}

We tested a limited number of models which may not fully capture the variability across different models and configurations. Furthermore, there is a potential risk that some dataset instances may overlap with the training data of the LLMs (especially for the Llama-3.1-8B model)  which could bias the evaluation results. Moreover, for the Zaytung dataset, the text field exceeded the sequence length of the masked language models (for BERT and RoBERTA). Therefore, we cropped the text fields for such instances.

It should also be noted that, using LLMs, specifically generative LLMs, ethical and environmental concerns should always be kept in mind. Generating textual data is an ethically convoluted topic, and should not be taken lightly. We believe that LLM-generated data should not be contextualized as if a real human has generated that content. These concerns may be limiting factors for the scalability of this study.

\section*{Acknowledgments}
The authors would like to thank Mustafa Ozan Alpay and Selin Özgürsoy for their hard work and contributions to data collection and annotation. 

This work is supported by the EuroHPC Development Access Call with Project ID DD-24-99. Parts of this research received the support of the EXA4MIND project, funded by the European Union's Horizon Europe Research and Innovation Programme, under Grant Agreement N° 101092944. Views and opinions expressed are however those of the author(s) only and do not necessarily reflect those of the European Union or the European Commission. Neither the European Union nor the granting authority can be held responsible for them.

\bibliography{custom}

\appendix
\section{Curating Human Annotations}
\label{sec:app_annotate}

As is, the satirical class of the \emph{Turkish Satirical News Dataset} is labelled as satirical since it is known to be collected from a satirical online newspaper. However, it is not analyzed to see what properties of the news articles make them satirical in the first place. 

To extend the usability fields of the curated dataset and obtain more information about the satirical corpus in the dataset as a baseline for explainability tasks, a subset of the \texttt{SATIRICAL} instances are further annotated by a human annotator. The annotation process is as follows:

\begin{enumerate}
    \item The main annotator goes through the whole article body and identifies the \texttt{REAL} and \texttt{FAKE} parts. 
    \item The \texttt{REAL} and \texttt{FAKE} markings are done according to the objective facts and events. The annotator is asked to fact-check and cite related information as needed.
    \item Four volunteers from different age demographics cross-check the annotations to have a higher coverage of news landscape knowledge.
    \item News articles with annotations that have a unified agreement are accepted, the rest is discarded.
\end{enumerate}

Finally, the human-annotated corpus consists of 40 satirical articles. Three selected annotations are shown in Figures \ref{fig:an1}, \ref{fig:an2}, and \ref{fig:an3}. The red text stands for the \texttt{FAKE} parts of the article, whereas the blue parts are marked as \texttt{REAL}.

\section{Comparing Model Explanations with Human Annotations}
\label{sec:app_shapselin}

\begin{figure*}[th!]
    \centering
    \begin{subfigure}[b]{0.45\textwidth}
        \centering
        \includegraphics[width=0.95\textwidth]{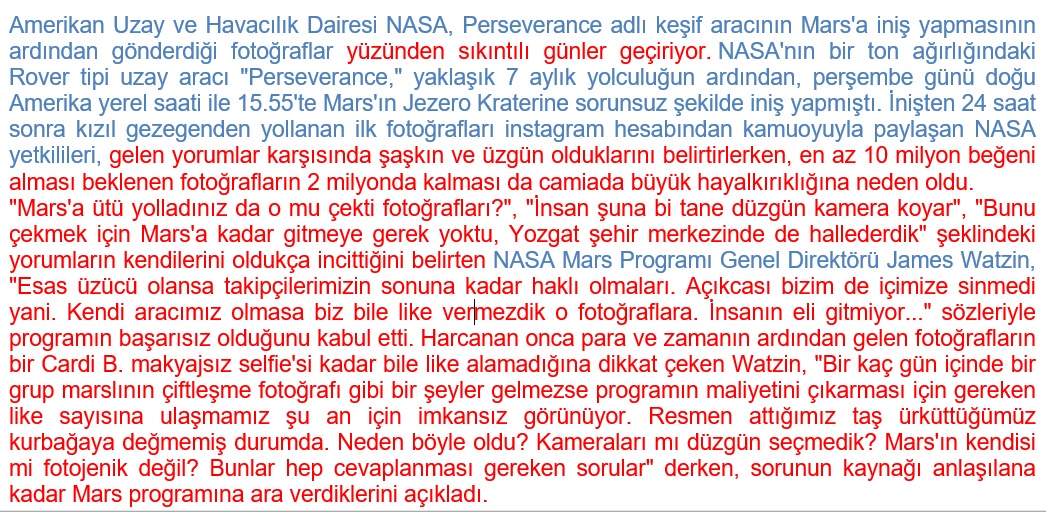}
  \caption{Human annotation for the article}
  \label{fig:an3}
    \end{subfigure}
    \hfill
    \begin{subfigure}[b]{0.45\textwidth}
        \centering
        \includegraphics[width=0.95\textwidth]{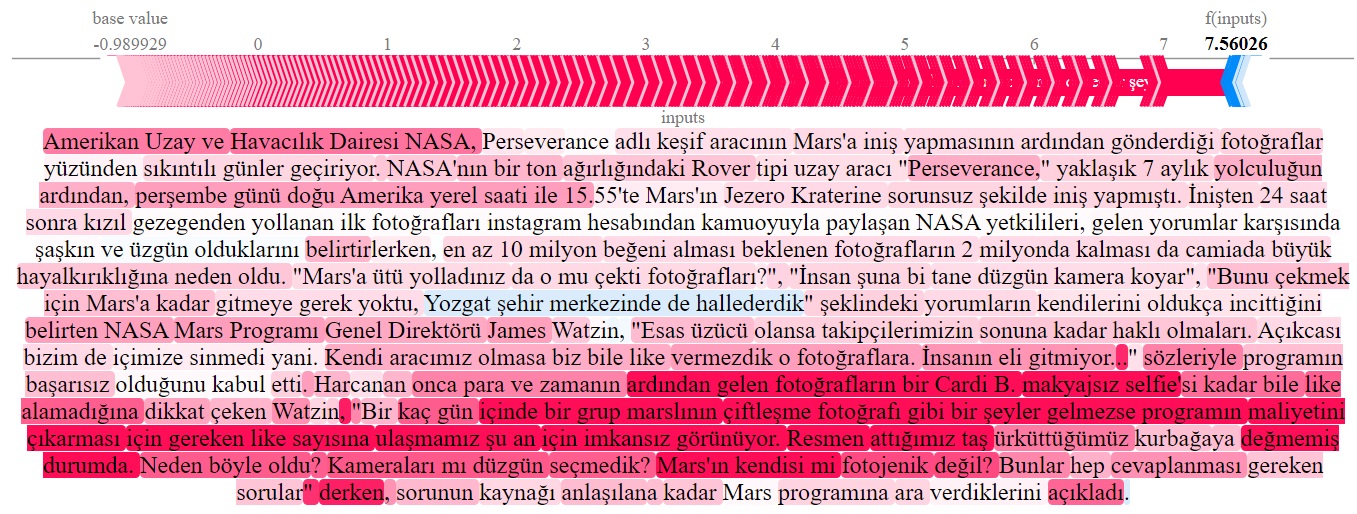}
  \caption{SHAP output for the article}
  \label{fig:sh3}
    \end{subfigure}
    \caption{Human annotation and SHAP annotation for Sample Article (1)}
    \label{fig:annotation3}
\end{figure*}

\begin{figure*}[th!]
    \centering
    \begin{subfigure}[b]{0.45\textwidth}
        \centering
        \includegraphics[width=0.95\textwidth]{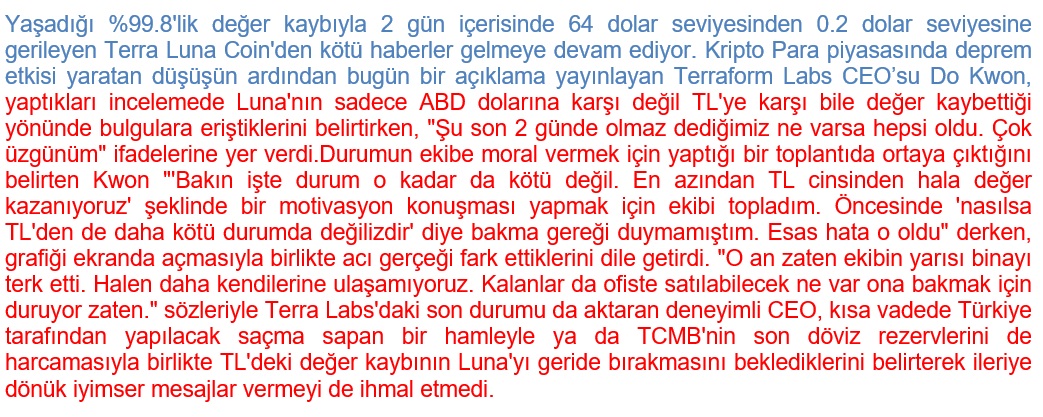}
  \caption{Human annotation for the article}
  \label{fig:an1}
    \end{subfigure}
    \hfill
    \begin{subfigure}[b]{0.45\textwidth}
        \centering
        \includegraphics[width=0.95\textwidth]{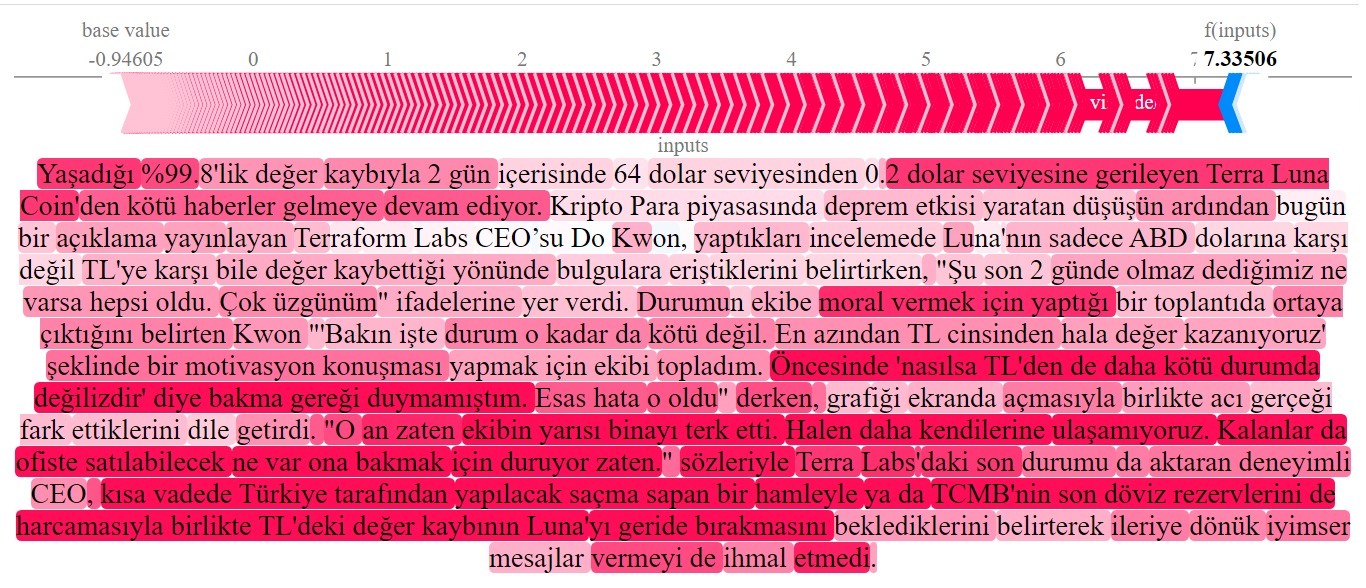}
  \caption{SHAP output for the article}
  \label{fig:sh1}
    \end{subfigure}
    \caption{Human annotation and SHAP annotation for Sample Article (2)}
    \label{fig:annotation1}
\end{figure*}

\begin{figure*}[th!]
    \centering
    \begin{subfigure}[b]{0.45\textwidth}
        \centering
        \includegraphics[width=0.95\textwidth]{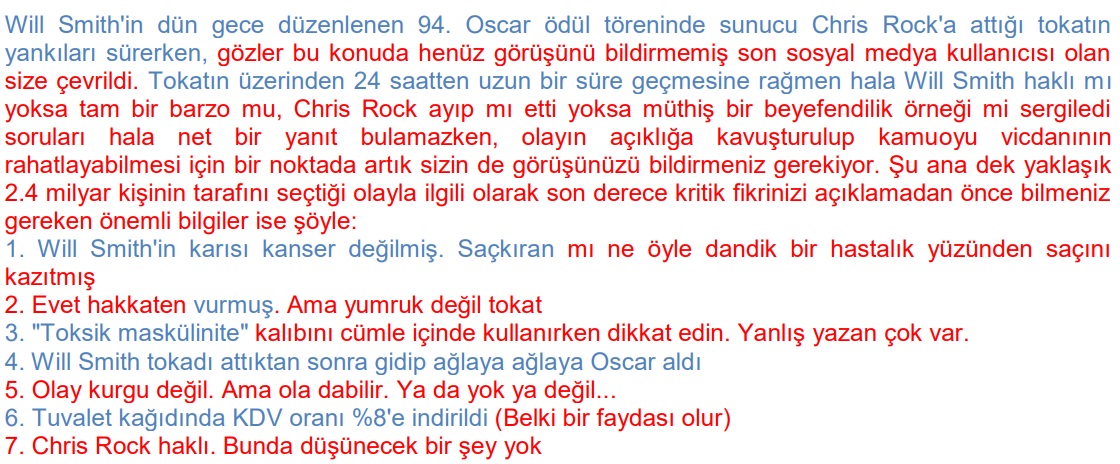}
  \caption{Human annotation for the article}
  \label{fig:an2}
    \end{subfigure}
    \hfill
    \begin{subfigure}[b]{0.45\textwidth}
        \centering
        \includegraphics[width=0.95\textwidth]{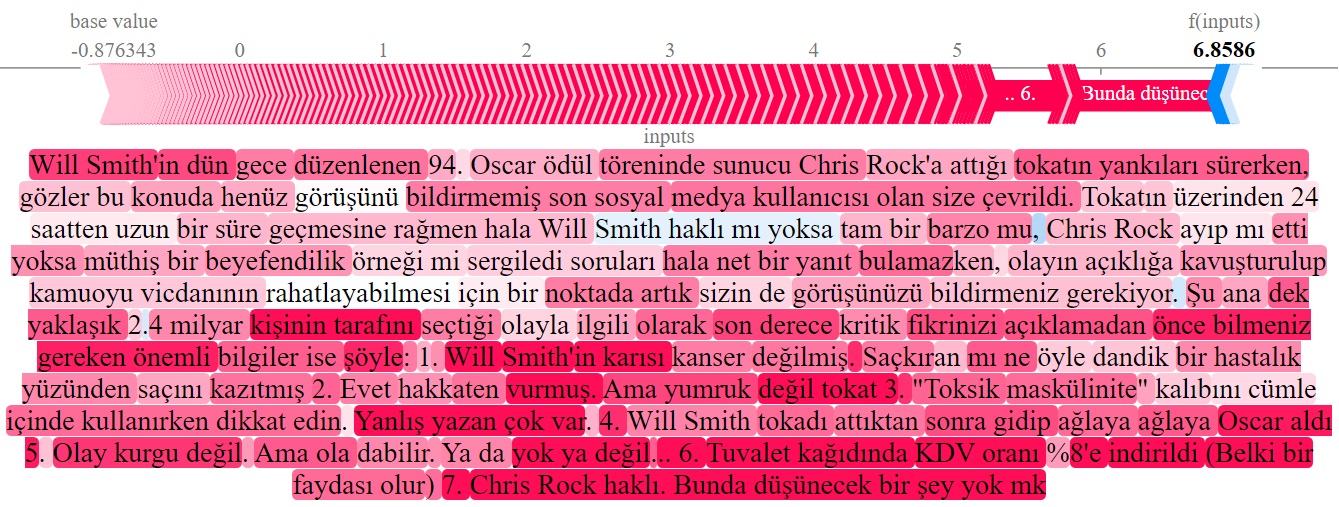}
  \caption{SHAP output for the article}
  \label{fig:sh2}
    \end{subfigure}
    \caption{Human annotation and SHAP annotation for Sample Article (3)}
    \label{fig:annotation2}
\end{figure*}

The curated \emph{Turkish Satirical News Dataset} includes human annotations for 40 \texttt{SATIRICAL} articles for utilizing explainable AI and interpretable ML methods on trained models. However, to draw a comparison between the human annotations and model explanations, it is needed to define a relation between satirical and fake content. Considering the nature of satirical news articles, it is assumed that the parts that are labeled as \texttt{FAKE} in the annotation are likely to contribute to the satirical meaning of the article. This can be in the form of a fake person, a fake quote, or a fake event. 

Similarly, the parts that are annotated to be \texttt{REAL} are less likely to contribute to the overall satire in the text. For example, the event described in an article may be real, therefore it can be annotated as \texttt{REAL}, but there may be a fake quote in the rest of the article that contributes to the satirical meaning. Following these parallels, a binary classifier is trained on \emph{Turkish Satirical News Dataset}, and the model decisions are explained using SHAP.

The SHAP \cite{shap} explainability method uses Shapley values to understand the relative importance of different features for a prediction instance of a model. In other words, it assigns importance values to the features relative to each other that show their weight in the final decision.

As a binary classifier model to be explained using SHAP, BERTurk \cite{stefan_schweter_2020_3770924} is fine-tuned. Later, SHAP-based explanations are extracted from the model. Three selected articles that have been classified correctly and have also been annotated by the human annotator are compared. One of these comparisons is reported in this section, and the other two are discussed in detail in Appendix \ref{sec:app_shapselin}, with English translations for all three articles.

The selected explanation is shown in Figure \ref{fig:sh1}, and its human-annotated counterpart is shown in Figure \ref{fig:an1}. The red highlights in the human annotation stand for the parts of the texts that are annotated as \texttt{FAKE} and the blue highlights specify the parts that are annotated as \texttt{REAL}. Similarly, for the SHAP output, red highlighted parts are explained as the \emph{important} parts of the texts that the classifier focuses on when identifying a data instance as \texttt{SATIRICAL}. Blue highlights in the SHAP output indicate that those parts of the texts are pulling the label towards \texttt{LEGITIMATE}, and the parts that are not highlighted are not important for the decision of the model. 

According to Figures \ref{fig:an1}, \ref{fig:sh1}, \ref{fig:an2}, \ref{fig:sh2}, \ref{fig:an3}, and \ref{fig:sh3} it can be seen that the SHAP output and the human annotation overlap for most of the red highlights, meaning that the expected match between \texttt{SATIRICAL} and \texttt{FAKE} annotations is observed. On the other hand, this seems not to be the case for blue highlights, i.e. for the parts that are annotated as \texttt{REAL} by the human annotators. It is observed that the model sometimes considers these parts as an indication of the \texttt{SATIRICAL} label or does not use those parts in the prediction at all. 

A closer look at article A (Figure \ref{fig:an1} and Figure \ref{fig:sh1}) shows that the red highlights for both the human annotation and model explanation generally match, but the blue highlights of the human annotator, i.e. the parts that are annotated as real correspond to the parts that are highlighted as slightly red or neutral by the model. Ideally, we would expect neutrality or blue highlights in the corresponding parts of the SHAP output. 

Additionally, a closer examination of article B (Figure \ref{fig:an2} and Figure \ref{fig:sh2}) shows that the model and human annotator are in disagreement for most of the annotations. Even though the more saturated reds highlighted by the model match the red highlights of the annotator, that is not the case for all of the red highlights. The SHAP output even shows blue highlights where the corresponding human annotation is red, or vice versa. 

Finally, a closer look at article C (Figure \ref{fig:an3} and Figure \ref{fig:sh3}) shows that both the human annotations and the SHAP output are less continuous than the previous two comparisons. Here, the highlights mostly line up with the human annotations, but the model generally misses the sudden truth value changes in a sentence, which is expected. 

Since the three examples all belong to the set of instances where the model predicts the class correctly, and the model already predicts almost every instance correctly, ideally it is expected to see a consistent overlap between the annotations and the SHAP explanations, which is not the case. This supports the claim that a model trained with a stylistically biased dataset may not be learning the representation of satire, but the style of the corpus instead.

Additionally, translations of Sample Articles (1) and (2) can be found in Appendix \ref{sec:app_prompts}. Translation for Sample Article (3) can be found in \ref{sec:app_others}.

\section{Prompts and Generated Articles in Turkish and English}
\label{sec:app_prompts}

Following two are the prompts in their original language, Turkish:

\begin{itemize}
    \item \textbf{(Prompt 1)} \emph{"Sana satirik bir haber vereceğim, adım adım bu haberdeki satirik unsurları kaldırmanı isteyeceğim. Önce bunun için haberden çıkarılması gereken cümleleri tespit et, sonra da cümleler çıkarılmış haliyle haberi tekrar yaz.\\Haber metni:"}
    \item \textbf{(Prompt 2)} \emph{"Sana bir metin vereceğim, içindeki satirik cümleleri daha düz bir dile çevirip tekrar yaz.\\Haber metni:"}
\end{itemize}

\subsection{Sample Article 1 generated with Prompt 1}

In Turkish:

\emph{"Amerikan Uzay ve Havacılık Dairesi NASA, Perseverance adlı keşif aracının Mars'a iniş yapmasının ardından gönderdiği fotoğraflar yüzünden sıkıntılı günler geçiriyor. Mars programının başarıyla sonuçlanan inişinden sonra, NASA yetkilileri beklenen ilginin aksine, sosyal medyada düşük bir etkileşimle karşılaştı. Program direktörü James Watzin, beklenen ilgiyi alamamalarının hayal kırıklığına yol açtığını ve programın başarısız olduğunu kabul etti. Sorunun kaynağını anlamak için Mars programına ara verildi."}

In English:

\emph{"The American Space and Aeronautics Agency NASA is going through difficult days due to the photos sent after the Perseverance rover successfully landed on Mars. Contrary to expectations, NASA officials encountered low engagement on social media following the successful landing of the Mars program. Program director James Watzin admitted that the lack of expected interest has led to disappointment and that the program has been deemed a failure. The Mars program has been paused to understand the source of the problem."}

\subsection{Sample Article 2 generated with Prompt 1 and Prompt 2}

Original article in Turkish:

Yaşadığı \%99.8'lik değer kaybıyla 2 gün içerisinde 64 dolar seviyesinden 0.2 dolar seviyesine gerileyen Terra Luna Coin'den kötü haberler gelmeye devam ediyor.
Kripto Para piyasasında deprem etkisi yaratan düşüşün ardından bugün bir açıklama yayınlayan Terraform Labs CEO’su Do Kwon, yaptıkları incelemede Luna'nın sadece ABD dolarına karşı değil TL'ye karşı bile değer kaybettiği yönünde bulgulara eriştiklerini belirtirken, "Şu son 2 günde olmaz dediğimiz ne varsa hepsi oldu. Çok üzgünüm" ifadelerine yer verdi.

Durumun ekibe moral vermek için yaptığı bir toplantıda ortaya çıktığını belirten Kwon "'Bakın işte durum o kadar da kötü değil. En azından TL cinsinden hala değer kazanıyoruz' şeklinde bir motivasyon konuşması yapmak için ekibi topladım. Öncesinde 'nasılsa TL'den de daha kötü durumda değilizdir' diye bakma gereği duymamıştım. Esas hata o oldu" derken, grafiği ekranda açmasıyla birlikte acı gerçeği fark ettiklerini dile getirdi.

"O an zaten ekibin yarısı binayı terk etti. Halen daha kendilerine ulaşamıyoruz. Kalanlar da ofiste satılabilecek ne var ona bakmak için duruyor zaten." sözleriyle Terra Labs'daki son durumu da aktaran deneyimli CEO, kısa vadede Türkiye tarafından yapılacak saçma sapan bir hamleyle ya da TCMB'nin son döviz rezervlerini de harcamasıyla birlikte TL'deki değer kaybının Luna'yı geride bırakmasını beklediklerini belirterek ileriye dönük iyimser mesajlar vermeyi de ihmal etmedi. 

Original article in English:

Following a 99.8\% loss in value, where Terra Luna Coin dropped from \$64 to \$0.20 within two days, bad news continues to emerge. After this dramatic drop that shook the cryptocurrency market, Do Kwon, CEO of Terraform Labs, released a statement today. He noted that their investigation revealed Luna had lost value not only against the US dollar but also against the Turkish Lira. Kwon said, "Everything we thought couldn’t happen in the last two days has happened. I am very sorry."

Kwon explained that the realization came during a meeting he held to boost the team’s morale, saying, “I gathered the team to give a motivational speech along the lines of ‘Look, it’s not so bad. At least we are still gaining value in TL terms.’ I hadn’t felt the need to check if we were worse off compared to TL. That was the real mistake.” He shared that the painful truth became apparent when they opened the graph on the screen.

“At that moment, half the team left the building. We still can’t reach them. The remaining team members are just looking around the office to see if there’s anything left to sell,” Kwon described the current situation at Terra Labs. The experienced CEO also mentioned that they expect the depreciation of the TL to eventually surpass Luna, whether due to some absurd move by Turkey or the Central Bank of the Republic of Turkey depleting its remaining foreign currency reserves. He added that he has not neglected to give optimistic messages for the future.

\subsubsection{Generated with Prompt 1}

In Turkish:

\emph{"Terra Luna Coin son günlerde büyük bir değer kaybı yaşadı. Terraform Labs CEO'su Do Kwon, yaptıkları incelemede Luna'nın sadece ABD dolarına karşı değil TL'ye karşı da değer kaybettiğini belirtti. Bu durum, şirket çalışanları arasında moral bozukluğuna neden oldu. Kwon, kısa vadede TL'deki değer kaybının Luna'yı geride bırakmasını beklediklerini ifade etti."}

In English:

\emph{"Terra Luna Coin has experienced a significant loss in value in recent days. Terraform Labs CEO Do Kwon stated that their investigation revealed that Luna has lost value not only against the US dollar but also against the Turkish Lira (TL). This situation has caused a morale decline among company employees. Kwon expressed that in the short term, they expect the depreciation of the TL to surpass that of Luna."}

\subsubsection{Generated with Prompt 2}

In Turkish:

\emph{"Son iki günde \%99.8 değer kaybı yaşayan Terra Luna Coin, 64 dolardan 0.2 dolara kadar geriledi ve kötü haberler gelmeye devam ediyor.\\
Bu büyük düşüşün ardından Terraform Labs CEO’su Do Kwon, bugün yaptığı açıklamada, Luna'nın sadece ABD dolarına değil, Türk Lirası'na karşı da değer kaybettiğini tespit ettiklerini söyledi. Kwon, "Bu son iki günde olmasını beklemediğimiz her şey oldu. Çok üzgünüm," dedi.\\
Kwon, durumu ekibe moral vermek için düzenlediği bir toplantıda fark ettiklerini belirtti. "Ekibe, 'En azından TL karşısında değer kazandık' demek istedim ama durumu kontrol etmemiştim. Asıl hata bu oldu," dedi. Grafiği açtıklarında ise acı gerçekle yüzleştiklerini ifade etti.\\
"Toplantı sırasında ekibin yarısı binayı terk etti ve hala onlara ulaşamıyoruz. Kalanlar ise ofiste satılacak bir şeyler arıyor," diye konuşan Kwon, kısa vadede TL'nin değer kaybının Luna'nınkini geride bırakabileceğini belirterek, geleceğe yönelik iyimser bir mesaj vermeyi de ihmal etmedi."}

In English:

\emph{"In the past two days, Terra Luna Coin has experienced a 99.8\% loss in value, dropping from \$64 to \$0.2, and bad news continues to emerge. \\
Following this major drop, Terraform Labs CEO Do Kwon stated in a recent announcement that they have found evidence that Luna has lost value not only against the US dollar but also against the Turkish Lira. Kwon said, "Everything we didn’t expect to happen in these last two days happened. I’m very sorry."\\
Kwon mentioned that they realized the situation during a meeting he organized to boost the team’s morale. "I wanted to tell the team, 'At least we’re gaining value against the TL,' but I hadn’t checked the situation. That was the real mistake," he said. He also mentioned that they faced the harsh reality when they opened the chart.\\
"During the meeting, half of the team left the building, and we still can't reach them. The rest are just looking for things to sell in the office," Kwon said. He also noted that in the short term, the loss of value of the TL might surpass that of Luna, and he did not forget to provide an optimistic message for the future."}

\subsection{Sample Article 4 generated with Prompt 2}

Original article in Turkish:

Bu sabah Taraf gazetesi tarafından ortaya atılan "Tutsaklara karşılık Süleyman Şah Türbesi IŞİD'a verilecek" şeklindeki şok haber, Dışişleri Bakanlığı tarafından kesin bir dille yalanlandı. Önce internet sitesinden yapılan açıklama sonra da Basın Sözcüsü Kamuran Aydilen aracılığı ile kamuoyunu aydınlatan Dışişleri Bakanlığı, "Ortadaki yanlış anlaşılmaları gidermek için söylüyoruz, Süleyman Şah Türbesi'nin yıkılarak yerine AVM yapılması konusunda IŞİD'le görüştüğümüz doğru. Neticede türbe yıkımında kendilerinden daha tecrübeli bir ekip yok. Ancak bunun dışında herhangi bir pazarlık söz konusu değil" ifadeleri ile iddiaları reddetti.

Bakanlık binasında gazetecilerin sorularını yanıtlayan Bakanlık Sözcüsü Aydilen, türbenin yıkım ihalesi için IŞİD ile pazarlık masasında oturulduğunu itiraf ederken, konunun rehinlerle doğrudan bir ilgisi bulunmadığını ise şu sözlerle savundu:

"Arkadaşlar 12 yıllık iktidarımızda artık bizi biraz tanımış olmanız lazım. Bütün dünya bilir ki biz, öyle 49 kişi için bir karış toprak vermeyiz. Hele de öyle bir toprağı, tam kupon arazi orası, deli misiniz ya? Mümkün mü böyle bir pazarlık? Türbeyi de geç, sırf arsası 4 milyar dolar eder. Orada nöbet tutan askerlerimize de sorduk, çevrede başka AVM de yokmuş. 'Çarşı izninde gidecek yer bulamıyoruz' diyorlar. Şu inşaat bir başlasın, Allah'ın izniyle para basacak orası..."

IŞİD'ın özellikle türbe yıkım işinde uzmanlaşmış, işlerini severek yapan ve sahiplenen bir örgüt olduğunun altını çizen Basın Sözcüsü, "Şu an bizden haber bekliyorlar, tamam dediğiimiz anda havanlarla falan girişecekler. Alimallah 1 saatte taş üstüne taş koymayız dediler. Rehineler konusunu öyle özel olarak konuşmadık ama o konuda bir jest yaparlarsa biz bunu geri çevirmeyiz elbette. Neticede birlikte iş yapan insanlarız, yarın öbür gün başka yıkım ihaleleri de olur... Bunları da değerlendireceklerdir" ifadelerine yer verdi. 

Mevcut anlaşmanın devletin kasasından bir kuruş çıkmadan halledileceğinin üzerinde duran Aydilen, yapılması planlanan AVM'nin detaylarını da basın mensuplarıyla paylaştı:

"Bakın buradan bööyle şimdiki türbenin kubbesi şeklinde bir tavan geliyor. Orası food court olacak... Alt katta SHAH'S SPORT adında bir fitness salonu ve atış poligonu var. Ta buraya kadar da meydan, forum mantığı gibi düşünün siz. Şimdi tabii aklınıza hemen ulaşım işi geliyor... Onu da düşündük. Hızlı treni 2017'de Marmaray'la Halkalı'ya bağladıktan sonra, Halkalı Ankara arası 4.5 saate inmiş olacak. Ankardan da ring seferiyle tak Halep'tesin. Son olarak Halep - Karakoza arası İDO'nun motorlarına binecek vatandaşlarımız anında AVM'de olacak. Bu kadar basit. Ayrıca oradaki askerlerimizi de özel güvenlik ve otopark görevlisi olarak AVM'de istihdam etmeyi düşünüyoruz. Gördüğünüz gibi bu projede kaybeden yok..." 

Bir soru üzerine Suleymanium AVM'yi, yaşasaydı Suleyman Şah'ın da takdirle karşılayacağını sözlerine ekleyen Dışişleri Sözcüsü, son olarak şunları kaydetti: 

"Yani düşünün tarihe geçmiş bir şahısınız, arkanızda bir tanecik kullanılmayan türbe kalıyor. Ne sineması var, ne otoparkı... Böyle mi anmalıyız ecdadımızı? Ayrıca son dönemde biliyorsunuz TOKİ'nin mevcut tarihi yapılar etrafında çeşitli çalışmaları mevcut. Sosyal medyada tarihi kümbetle iç içe geçmiş yurtlarımız büyük ilgi gördü. Bu şekilde alışveriş keyfini manevi iklimle birleştiren bir çalışma halkımızın da ilgisini çekecektir..." 

Original article in English:

This morning, the shocking news reported by Taraf newspaper, claiming "The Süleyman Shah Tomb will be handed over to ISIS in exchange for the hostages," was firmly denied by the Ministry of Foreign Affairs. The Ministry, which clarified the issue first through an announcement on its website and later through its Spokesperson Kamuran Aydilen, stated: "To clear up the misunderstandings, we are saying this: it is true that we have discussed with ISIS the demolition of the Süleyman Shah Tomb and replacing it with a shopping mall. After all, there is no team more experienced than them when it comes to demolitions. However, there is no bargaining or negotiation beyond this."

While answering journalists' questions at the Ministry building, Spokesperson Aydilen admitted that negotiations with ISIS had taken place regarding the demolition of the tomb, but he defended that the issue had nothing to do with the hostages, saying:

"Friends, you should have known us by now after 12 years in power. The whole world knows that we wouldn’t give up an inch of land for 49 people. Especially not such a prime piece of land, are you crazy? How could there be such a deal? Forget the tomb, just the land itself is worth 4 billion dollars. We even asked the soldiers stationed there, and they said there aren’t any other shopping malls around. They say, 'We can’t find a place to go on our leave.' Once construction starts, God willing, that place will be printing money..."

Emphasizing that ISIS is a group specialized in demolitions, especially of tombs, and one that enjoys and takes ownership of their work, the Spokesperson added: "Right now, they are waiting for our confirmation. As soon as we say 'yes,' they’ll come in with mortars and the like. I swear, they said they’ll finish the job in an hour. We didn’t specifically discuss the hostages, but if they make a gesture in that regard, we certainly won’t turn it down. After all, we’re people who work together, and who knows, there may be other demolition tenders in the future… They’ll certainly evaluate those as well."

Aydilen stressed that the current agreement would be handled without spending a penny from the state’s coffers and shared details of the planned shopping mall with the press:

"Look, we’re bringing in a ceiling shaped like the current dome of the tomb. That area will be a food court... On the lower floor, there will be a fitness center called SHAH'S SPORT and a shooting range. Think of it like a forum or square. Now, of course, you’re probably thinking about transportation... We’ve thought of that too. After Marmaray connects to Halkalı in 2017, the travel time from Halkalı to Ankara will be reduced to 4.5 hours. From Ankara, a shuttle service will quickly take you to Aleppo. Lastly, our citizens traveling between Aleppo and Karakoza will take the IDO ferries and will be at the shopping mall in no time. It’s that simple. We also plan to employ our soldiers stationed there as security and parking attendants in the shopping mall. As you can see, no one loses in this project..."

In response to a question, the Ministry Spokesperson added that the Süleymanium Shopping Mall would have been approved by Süleyman Shah himself, had he been alive:

"Imagine, you’re a historical figure, and the only thing left behind is an unused tomb. It has no cinema, no parking lot... Is this how we should remember our ancestors? Also, as you know, TOKİ (Housing Development Administration of Turkey) has been working around existing historical structures. Our dormitories, integrated with the historical domes, have garnered a lot of attention on social media. This project, which combines the joy of shopping with a spiritual atmosphere, will also attract the public’s interest..."

Generated article in Turkish:

\emph{"Taraf gazetesinin "Tutsaklara karşılık Süleyman Şah Türbesi IŞİD'e verilecek" iddiası Dışişleri Bakanlığı tarafından kesin bir dille yalanlandı. Bakanlık, Süleyman Şah Türbesi'nin yıkılarak yerine AVM yapılması konusunda IŞİD'le görüştüklerini, ancak bu görüşmenin rehinelerle ilgisi olmadığını belirtti.\\
Bakanlık Sözcüsü Kamuran Aydilen, "IŞİD yıkım işinde deneyimli bir ekip. Türbeyi yıkmak için onlarla görüşüyoruz ama toprak verme pazarlığı yok. Türbenin arsası çok değerli, rehineler için böyle bir pazarlık olmaz. Orada AVM yaparak ekonomiye katkı sağlamayı planlıyoruz" dedi.\\
Aydilen, yapılacak AVM'nin türbenin kubbesi şeklinde bir tavanı olacağını, food court, fitness salonu ve atış poligonu gibi bölümler içereceğini, ulaşımın hızlı tren ve İDO motorları ile sağlanacağını açıkladı. Askerlerin de AVM'de özel güvenlik ve otopark görevlisi olarak istihdam edileceğini ekledi. Ayrıca, bu projeyle kimsenin kaybetmeyeceğini ve halkın ilgisini çekeceğini belirtti."}

Generated article in English:

\emph{"The claim made by Taraf newspaper that the Süleyman Shah Tomb will be handed over to ISIS in exchange for hostages has been strongly denied by the Ministry of Foreign Affairs. The ministry clarified that they are in talks with ISIS regarding the demolition of the tomb to build a shopping mall in its place, but emphasized that these discussions are unrelated to the hostages.\\
Ministry spokesperson Kamuran Aydilen stated, "ISIS has an experienced demolition team. We're negotiating with them to demolish the tomb, but there is no land-for-hostages deal. The land where the tomb is located is very valuable, and we wouldn’t negotiate it for hostages. We plan to contribute to the economy by building a shopping mall there."\\
Aydilen also mentioned that the mall will have a dome-shaped ceiling inspired by the tomb's dome and will include sections such as a food court, fitness center, and shooting range. Transportation to the mall will be provided by high-speed trains and İDO ferries. He added that the soldiers stationed there will be employed as security personnel and parking attendants at the mall. Furthermore, he emphasized that this project would not cause any losses and would attract public interest."}

\section{Other articles}
\label{sec:app_others}

\subsection{Translation of Sample Article (3)}

Warning: You're the Only One Who Hasn't Weighed In on Will Smith's Slap...

As the repercussions of Will Smith’s slap on comedian Chris Rock at the 94th Oscars last night continue to ripple, all eyes have turned to you, the last social media user who has yet to express an opinion on the matter. Over 24 hours have passed since the slap, and the questions of whether Will Smith was justified or simply out of line, and whether Chris Rock was rude or demonstrated exemplary gentlemanliness, still lack clear answers. For the sake of clarifying the situation and easing the public conscience, it’s time for you to share your view.

Before you disclose your crucial opinion on this event, which has involved approximately 2.4 billion people choosing sides, here are some important details you need to know:

1. Will Smith's wife is not suffering from cancer. She shaved her head due to some trivial disease like alopecia.
2. Yes, he really did hit him. But it was a slap, not a punch.
3. Be careful when using the term "toxic masculinity" in a sentence. Many people spell it wrong.
4. After delivering the slap, Will Smith went on to cry and then won the Oscar.
5. The incident is not staged. But it could be. Or not... who knows.
6. The VAT rate on toilet paper has been reduced to 8\% (Maybe this will help).
7. Chris Rock is right. There’s nothing more to think about.

\end{document}